\documentclass{article}

\def\vec#1{\mathbf{#1}}

\usepackage[preprint,nonatbib]{nips_2018}

\usepackage[dvipdf]{graphicx}
\usepackage[utf8]{inputenc} 
\usepackage[T1]{fontenc}    
\usepackage{hyperref}       
\usepackage{url}            
\usepackage{booktabs}       
\usepackage{amsfonts}       
\usepackage{microtype}      

\title{Unsupervised Object Matching for Relational Data}

\author{
  Tomoharu~Iwata
  \hspace{3em}
  Naonori~Ueda\\
  NTT Communication Science Laboratories\\  
  \texttt{\{iwata.tomoharu,ueda.naonori\}@lab.ntt.co.jp}
}

\usepackage{amsmath}	
\begin{document}

\maketitle

\begin{abstract}
We propose an unsupervised object matching method for relational data, which finds matchings between objects in different relational datasets without correspondence information. For example, the proposed method matches documents in different languages in multi-lingual document-word networks without dictionaries nor alignment information. The proposed method assumes that each object has latent vectors, and the probability of neighbor objects is modeled by the inner-product of the latent vectors, where the neighbors are generated by short random walks over the relations. The latent vectors are estimated by maximizing the likelihood of the neighbors for each dataset. The estimated latent vectors contain hidden structural information of each object in the given relational dataset. Then, the proposed method linearly projects the latent vectors for all the datasets onto a common latent space shared across all datasets by matching the distributions while preserving the structural information. The projection matrix is estimated by minimizing the distance between the latent vector distributions with an orthogonality regularizer. To represent the distributions effectively, we use the kernel embedding of distributions that hold high-order moment information about a distribution as an element in a reproducing kernel Hilbert space, which enables us to calculate the distance between the distributions without density estimation. The structural information encoded in the latent vectors are preserved by using the orthogonality regularizer. We demonstrate the effectiveness of the proposed method with experiments using real-world multi-lingual document-word relational datasets and multiple user-item relational datasets.
\end{abstract}

\section{Introduction}

A wide variety of data are represented as relational data, 
such as
friend links on social networks,
hyperlinks between web pages,
citations between scientific articles,
word appearance in documents,
item purchase by users,
and interactions between proteins.
In this paper, we consider object matching for relational data,
which is the task to find correspondence between objects
in different relational datasets.
For example, corresponding objects are
words with the same meaning in different languages~\cite{haghighi2008},
the same users in different databases~\cite{korula2014efficient},
related entities in different
ontologies~\cite{doan2002learning,aumueller2005schema},
and proteins with the same function in different species~\cite{kuchaiev2010topological}.
There have been proposed many methods for object matching
when a similarity measure is defined between objects in different domains
or correspondence information is given~\cite{socher2010connecting,zhou2012factorized,terada2012}.
However, 
similarity measures might not be defined across different modalities,
and correspondence information would be unavailable 
due to the need to preserve privacy or its high cost.

\begin{figure}
\centering
\includegraphics[height=10em]{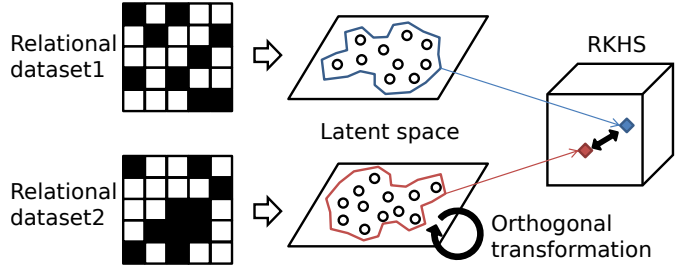}
\caption{The proposed unsupervised object method for relational datasets. The proposed method 1) obtains latent vectors for objects that contain structural information for each dataset, 2) embeds the distributions of the latent vectors in a RKHS, and 3) finds an orthogonal transformation by minimizing the distance between the distributions in the RKHS.}
\label{fig:mmd_matching}
\end{figure}

We propose an unsupervised method to find matchings of objects in different relational datasets
without similarity measures nor correspondence information.
First, the proposed method estimates 
latent vectors for objects 
by making use of representation learning on graphs~\cite{perozzi2014deepwalk}
for each relational dataset.
In the latent vectors, hidden structural information about the relational dataset is encoded
by modeling neighbor objects with the inner-product of the latent vectors,
where the neighbors are generated by short random walks 
over the relations.
Then, the latent vectors are linearly projected
onto a common latent space shared across all relational datasets
by matching the latent vector distributions
while preserving the encoded structural information.
To represent the distributions effectively,
we use the kernel embeddings of distributions~\cite{sriperumbudur2010hilbert},
that hold high-order moment information about a distribution
as an element in a reproducing kernel Hilbert space (RKHS).
It enables us to calculate the distance between
the distributions, which
is called maximum mean discrepancy (MMD)~\cite{gretton2012kernel}, 
without density estimation.
The structural information is preserved by using 
an orthogonal projection matrix 
since it does not change the values of the inner-product.
We estimate a projection matrix by minimizing the MMD between
the latent vector distributions of different relational datasets
with an orthogonality regularizer.
Objects to be matched are estimated based
on the distance in the common latent space.
Figure~\ref{fig:mmd_matching} shows an overview of the proposed method.


\section{Related work}
\label{sec:related}

A number of unsupervised object matching methods have been
proposed~\cite{haghighi2008,quadrianto2010,djuric2012convex,klami2012,klami2013bayesian,yamada2011cross}, such as kernelized sorting~\cite{quadrianto2010} 
and Bayesian object matching~\cite{klami2013bayesian}.
However, these methods are not for relational data.
In addition, these methods do not scale to large data
since they find correspondence by estimating
a (probabilistic) permutation matrix with the size of the square of the number of objects.
On the other hand, the proposed method scales well
since it estimates a projection matrix with the size of the square of
the latent space dimensionality, which is much smaller than the number of objects.

The ReMatch method is an unsupervised cluster matching method 
for relational data~\cite{iwata2016}.
The ReMatch assigns each object into a cluster 
that is shared across all datasets,
and finds correspondence based on the cluster assignments.
Therefore, multiple objects assigned into the same cluster
cannot be distinguished, and
there would be many ties when objects are ranked based on the estimated correspondence.
In contrast, the proposed method estimates different continuous feature representations for different objects.

In natural language processing, 
methods for word translation without parallel data
have been proposed~\cite{cao2016distribution,zhang2017earth,zhang2017adversarial,lample2018word}.
With these methods, 
word embeddings obtained by word2vec~\cite{mikolov2013distributed}
are transformed by matching the distributions
with the orthogonality constraint~\cite{xing2015normalized,smith2017offline}.
Although the proposed method employs a similar approach with these methods,
there are two clear differences.
The first difference is that the proposed method is for relational data,
but these methods are for natural language sentences.
The second difference is techniques for matching distributions.
The distributions are matched based their mean and variance in~\cite{cao2016distribution},
which implicitly assumes the latent vectors follow Gaussian distributions. 
The proposed method uses the kernel embeddings of distributions, 
by which higher-order moments as well as mean and covariance
are considered for matching.
The latent vector distributions of relational data
are generally not Gaussians as shown in Figure~\ref{fig:vis} in our experiments.
The earth mover's distance is used for the distribution distance measure in~\cite{zhang2017earth},
which requires solving a minimization problem.
In contrast, the MMD used with the proposed method 
is calculated in a closed form by the weighted sum of the kernel functions
without the need of optimization.
Adversarial training that solves a minimax optimization problem
is used for matching the distributions in~\cite{zhang2017adversarial,lample2018word}.
Although the adversarial training is successfully used especially
for image generation~\cite{goodfellow2014generative},
it is well known for being unstable~\cite{arjovsky2017towards},
and requires training an additional neural network for a discriminator.
The MMD has fewer tuning hyperparameters than the adversarial training approach,
and the proposed method works stably as empirically shown in Section~\ref{sec:experiments}.

The kernel embeddings of distributions and the MMD
have been successfully used for various applications, such as
a statistical independence test~\cite{gretton2008kernel,NIPS2012_4727,zaremba2013b},
discriminative learning on probability measures~\cite{muandet2012learning},
anomaly detection for group data~\cite{muandet2013one},
density estimation~\cite{dudik2007maximum},
a three-variable interaction test~\cite{sejdinovic2013kernel},
a goodness of fit test~\cite{chwialkowski2016kernel},
supervised object matching~\cite{yoshikawa2015cross},
domain adaptation~\cite{long2015learning},
and deep generative modeling~\cite{li2015generative,li2017mmd}.

The proposed method is related to graph matching and network alignment methods,
which find correspondence so that matched objects
have the same edge relation and/or the similar attributes~\cite{NIPS2006_2960,NIPS2009_3800,NIPS2009_3756,NIPS2013_4925,NIPS2017_6911,kuchaiev2010topological,terada2012,cho2010reweighted,gori2005exact,sharma2010shape}.
The proposed method is different from them, in a way that
the proposed method considers not only directly connected edge relation but also
the intrinsic hidden structure of the given relational dataset without attributes
by embedding objects onto a latent space.
The proposed method is based on the recent advancements of graph representation 
learning methods, such as DeepWalk~\cite{perozzi2014deepwalk}, LINE~\cite{tang2015line}
and node2vec~\cite{grover2016node2vec}.
These methods obtain node representations 
that contain structural information in the graph,
and have been successfully used in link prediction and node classification.
Although spectral methods for graph matching have been proposed~\cite{knossow2009inexact,umeyama1988eigendecomposition}, DeepWalk and node2vec achieved better performance than the spectral methods in graph representation learning~\cite{perozzi2014deepwalk,grover2016node2vec}.

\section{Kernel embeddings of distributions}
\label{sec:KED}

In this section,
we explain the kernel embeddings of distributions, 
which are employed with the proposed method.
The kernel embeddings of distributions~\cite{sriperumbudur2010hilbert} 
embed probability distribution $\mathbb{P}$ on space $\mathcal{X}$ into a reproducing kernel Hilbert space (RKHS) $\mathcal{H}_k$ specified by kernel $k$.
In particular, distribution $\mathbb{P}$ is 
represented as an element $\mu^*(\mathbb{P})$ in the RKHS as follows, 
$\mu^{*}(\mathbb{P}) = \mathbb{E}_{\vec{x} \sim \mathbb{P}}[k(\cdot, \vec{x})]
= \int_\mathcal{X} k(\cdot, \vec{x}) d\mathbb{P} \in \mathcal{H}_k$. 
The kernel embedding $\mu^{*}(\mathbb{P})$ preserves the properties of the distribution $\mathbb{P}$ such as mean, covariance and higher-order moments when characteristic kernels, such as Gaussian kernels, are used~\cite{sriperumbudur2010hilbert}.

Although distribution $\mathbb{P}$ is unknown in practice, 
we can estimate the empirical kernel embedding using a set of samples
$\vec{X} = \{\vec{x}_{n} \}_{n=1}^{N}$ drawn from the distribution.
By interpreting the samples $\vec{X}$ as an empirical distribution 
$\hat{\mathbb{P}} = \frac{1}{N}\sum_{n=1}^{N} \delta_{\vec{x}_{n}}(\cdot)$, 
where $\delta_\vec{x}(\cdot)$ is the Dirac delta function, 
empirical kernel embedding $\mu(\vec{X})$ is given by
$\mu(\vec{X}) = \frac{1}{N} \sum_{n=1}^{N} k(\cdot,\vec{x}_{n}) \in \mathcal{H}_k$,
which can be approximated with an error rate of $\parallel\mu(\vec{X})-\mu^*(\mathbb{P})\parallel_{\mathcal{H}_k}=O_p(N^{-\frac{1}{2}})$~\cite{smola2007hilbert}.
Unlike kernel density estimation, the error rate of the kernel embeddings is independent of the dimensionality of the given distribution.

By using the kernel embeddings, we can measure the distance between two distributions.
Given two set of vectors, $\vec{X} = \{\vec{x}_{n}\}_{n=1}^{N}$ and $\vec{Y}=\{\vec{y}_{m}\}_{m=1}^{M}$, 
we obtain their kernel embeddings $\mu(\vec{X})$ and $\mu(\vec{Y})$.
Then, the distance between $\mu(\vec{X})$ and $\mu(\vec{Y})$ is given by 
\begin{align}
{\rm MMD}^{2}(\vec{X}, \vec{Y}) 
= \parallel \mu(\vec{X})\!-\!\mu(\vec{Y}) \parallel^2_{\mathcal{H}_k}
= \langle \mu(\vec{X}), \mu(\vec{X}) \rangle_{\mathcal{H}_k}
\!+\! \langle \mu(\vec{Y}), \mu(\vec{Y}) \rangle_{\mathcal{H}_k} 
\!-\! 2 \langle \mu(\vec{X}), \mu(\vec{Y}) \rangle_{\mathcal{H}_k},
\label{eq:MMD}
\end{align}
where $\langle \cdot, \cdot \rangle_{\mathcal{H}_k}$ is an inner-product in the RKHS, and this distance is called 
maximum mean discrepancy (MMD)~\cite{gretton2008kernel}.
The inner-product is calculated by
\begin{align}
\langle \mu(\vec{X}), \mu(\vec{Y}) \rangle_{\mathcal{H}_k}
= \left\langle \frac{1}{N} \sum_{i=1}^{N} k(\cdot, \vec{x}_{i}), \frac{1}{M} \sum_{j=1}^{M} k(\cdot, \vec{y}_{j})  \right\rangle_{\mathcal{H}_k}
= \frac{1}{NM} \sum_{i=1}^{N} \sum_{j=1}^{M} k(\vec{x}_i, \vec{y}_j),
\label{eq:inner}
\end{align}
which is represented by an empirical expectation of the kernel.
When the number of vectors is large, the computation can be expensive.
In that case, we can obtain the unbiased estimate of the MMD
by using Monte Carlo approximation 
for the expectation~\cite{NIPS2012_4727,long2015learning},
in which vectors are randomly sampled to compute the expectation.

\section{Proposed method}
\label{sec:proposed}

\subsection{Task}
\label{sec:task}

Suppose that we are given $D$ relational datasets, or networks,
$\{{\cal G}_{d}\}_{d=1}^{D}$.
Here, ${\cal G}_{d}=({\cal V}_{d},{\cal E}_{d})$ is the $d$th relational dataset,
${\cal V}_{d}=\{v_{d1},\cdots,v_{dN_{d}}\}$ is the set of objects,
$N_{d}$ is the number of objects, 
and ${\cal E}_{d}$ is the set of relations.
The proposed method is applicable to arbitrary kinds of relational data,
such as single-type, multi-type and/or weighted relations,
if all the given datasets are the same type.
The task is to find correspondence between objects 
in different relational datasets.

\subsection{Procedures}
\label{sec:procedures}

\paragraph{Latent vector estimation}
With the proposed method, 
continuous feature representations for objects are 
obtained in each relational dataset.
We obtain the object representations
by using a skip-gram~\cite{mikolov2013efficient} 
based approach for graphs~\cite{perozzi2014deepwalk,grover2016node2vec}.
This approach encodes structural information of the given graph
as continuous vectors.
In particular, each object $v_{dn}$ is assumed to have
two latent vectors $\vec{u}_{dn}\in\mathbb{R}^{K}$ and $\vec{h}_{dn}\in\mathbb{R}^{K}$, where $K$ is the dimensionality of the latent space.
The probability that object $v_{dm}$ is the neighbor of object $v_{dn}$
is modeled by the inner-product of the latent vectors as follows,
\begin{align}
p(v_{dm}|v_{dn}) = \frac{\exp(\vec{u}_{dm}^{\top}\vec{h}_{dn})}
 {\sum_{m'=1}^{N_{d}}\exp(\vec{u}_{dm'}^{\top}\vec{h}_{dn})}.
\label{eq:neighbor_prob}
\end{align}
The latent vectors for all objects in the $d$th dataset,
$\vec{U}_{d}=\{\vec{u}_{dn}\}_{n=1}^{N_{d}}$ and
$\vec{H}_{d}=\{\vec{h}_{dn}\}_{n=1}^{N_{d}}$,
are obtained by maximizing the following likelihood of neighbors,
\begin{align}
L_{d}(\vec{U}_{d},\vec{H}_{d}) = \sum_{n=1}^{N_{d}} \sum_{v_{dm}\in{\cal N}(v_{dn})}\log p(v_{dm}|v_{dn}),
\label{eq:likelihood}
\end{align}
where ${\cal N}(v_{dn})\subseteq {\cal V}_{d}$ is the set of neighbors of object $v_{dn}$.
The summation over all the objects in the denominator of 
(\ref{eq:neighbor_prob})
is expensive to compute for large data.
We approximate it with negative sampling~\cite{mikolov2013distributed}
for efficiency.

The neighbors are generated by short random walks over relations ${\cal E}_{d}$.
We conduct multiple random walks starting from every objects.
A random walk chooses a next object uniform randomly
from the objects that have relations with the current object
until the maximum length is reached.
The objects that appear in the random walk sequences within a window
are considered as the neighbors.
The random walks are successfully used for capturing structure
in graphs~\cite{andersen2006local},
and robust to perturbations in the form of
noisy or missing relations~\cite{grover2016node2vec}.
The neighbors are not restricted to objects directly connected by relations,
but are generated depending on the local relational structure.

\paragraph{Projection matrix estimation}
Since we obtained object representations independently for each dataset,
the obtained representations, $\vec{U}_{d}$ and $\vec{H}_{d}$, 
are not related across different datasets.
Then, we project the representations 
onto a common latent space shared across all datasets
by matching the distributions 
while preserving the encoded structural information.
We assume the following linear transformation with orthogonal projection matrix $\vec{W}_{d}\in\mathbb{R}^{K\times K}$,
\begin{align}
\tilde{\vec{u}}_{dn}=\vec{W}_{d}\vec{u}_{dn},\quad
\tilde{\vec{h}}_{dn}=\vec{W}_{d}\vec{h}_{dn},
\end{align}
where $\tilde{\vec{u}}_{dn}\in\mathbb{R}^{K}$ and
$\tilde{\vec{h}}_{dn}\in\mathbb{R}^{K}$ are the transformed latent vector of 
$\vec{u}_{dn}$ and $\vec{h}_{dn}$, respectively.
Here, without loss of generality 
we can set that the transformation for the first dataset
is the identity matrix, $\vec{W}_{1}=\vec{I}$.
By assuming the orthogonality $\vec{W}_{d}^{\top}\vec{W}_{d}=\vec{I}$, 
we can preserve the encoded structural information,
since the inner-product of the transformed vectors is the same with the
that of the original vectors,
$\tilde{\vec{u}}_{dn}^\top\tilde{\vec{h}}_{dm}
=\vec{u}_{dn}^{\top}\vec{W}_{d}^{\top}\vec{W}_{d}\vec{h}_{dm}
=\vec{u}_{dn}^{\top}\vec{h}_{dm}$,
and the relations are modeled
with only the inner-product in the proposed method
in (\ref{eq:neighbor_prob}). 

We would like to have the transformed vectors that follows 
the same distribution among all datasets. 
We employ the kernel embeddings of distributions 
to represent the distribution of the transformed latent vectors of the $d$th dataset
as follows,
$\mu(\tilde{\vec{U}}_{d})=\frac{1}{N_{d}}\sum_{n=1}^{N_{d}}k(\cdot,\tilde{\vec{u}}_{dn})$, and
$\mu(\tilde{\vec{H}}_{d})=\frac{1}{N_{d}}\sum_{n=1}^{N_{d}}k(\cdot,\tilde{\vec{h}}_{dn})$.
Then, the distance between the latent vector distributions of datasets $d$ and $d'$ is measured by
${\rm MMD}^{2}(\tilde{\vec{U}}_{d},\tilde{\vec{U}}_{d'})=\parallel \mu(\tilde{\vec{U}}_{d})-\mu(\tilde{\vec{U}}_{d'})\parallel_{{\cal H}_{k}}^{2}$,
and
${\rm MMD}^{2}(\tilde{\vec{H}}_{d},\tilde{\vec{H}}_{d'})=\parallel \mu(\tilde{\vec{H}}_{d})-\mu(\tilde{\vec{H}}_{d'})\parallel_{{\cal H}_{k}}^{2}$,
which are calculated using (\ref{eq:MMD}) and (\ref{eq:inner}).

We obtain orthogonal matrices $\vec{W}=\{\vec{W}_{d}\}_{d=2}^{D}$,
by which transformed vectors follow the same distribution,
by minimizing the following objective function,
\begin{align}
E(\vec{W}) = \sum_{d=2}^{D}\parallel\vec{W}_{d}^{\top}\vec{W}_{d}-\vec{I}\parallel^{2}
+ 
\lambda \sum_{d=1}^{D}\sum_{d'=d+1}^{D}
\Bigl(
{\rm MMD}^{2}(\tilde{\vec{U}}_{d},\tilde{\vec{U}}_{d'})+{\rm MMD}^{2}(\tilde{\vec{H}}_{d},\tilde{\vec{H}}_{d'})
\Bigr),
\label{eq:loss}
\end{align}
where the first term handles the orthogonality,
the second term handles the distribution matching,
and the $\lambda\in\mathbb{R}_{\geq 0}$ is the hyperparameter.
The objective function is minimized
by using gradient based optimization methods.

\paragraph{Matching}
The object correspondence is calculated based on the Euclidean distance
of the transformed latent vectors $\{\tilde{\vec{U}}_{d}\}_{d=1}^{D}$
in the common latent space.
The ranking of objects in the $d'$th dataset to be matched 
with object $v_{dn}$ is obtained by the
ascending order of the Euclidean distance
$\parallel \tilde{\vec{u}}_{dn}-\tilde{\vec{u}}_{d'n'}\parallel$.

\subsection{Model}
\label{sec:model}

We described the procedures for matching with the proposed method in the
previous subsection. In this subsection, we explain the model the
proposed method assumes.

The proposed model assumes that each object $v_{dn}$ 
in all datasets has latent
vectors $\tilde{\vec{u}}_{dn}$ and $\tilde{\vec{h}}_{dn}$ 
in a common latent space. The distance in the common latent
space indicates the correspondence, and the objects that are closely
located to each other are considered to be matched. The latent vectors in
all datasets are generated from common distributions, 
$p(\tilde{\vec{u}})$ and $p(\tilde{\vec{h}})$. 
The proposed method does not explicitly model the common distributions. 
Instead, we achieve to have the common distributions 
by minimizing the distributions across different datasets based on MMD. 
By this approach, we do not need to consider 
the parametric form of the distributions, 
and do not need to estimate the distributions.

The neighbors, which are defined by the relations, 
are assumed to be modeled using the latent vectors 
with (\ref{eq:neighbor_prob}).
The neighbor model (\ref{eq:neighbor_prob}) 
does not contain dataset-specific parameters except for the latent vectors, 
and the form of the model is the
same across all datasets given the latent vectors. 
By this modeling, 
all we need to consider is the latent vector distributions
for finding correspondence.

Although we can estimate the latent vectors in the common space $\{\tilde{\vec{U}}_{d}\}_{d=1}^{D}$ and $\{\tilde{\vec{H}}_{d}\}_{d=1}^{D}$
directly in a one-step approach, the proposed method employs 
the following two-step approach: 
1) estimates individual latent vectors for each dataset, and
then transforms them into the common space. The two-step approach has
advantages over the one-step approach. First, the two-step approach is
easy to parallelize the latent vector estimation. 
Second, we can estimate the latent vectors robustly by separating 
the structural information encoding and distribution matching. 
On the other hand, the one-step approach can deteriorate 
the encoding quality by enforcing the distribution matching. 
In our preliminary experiments, the two-step approach
performed better than the one-step approach. 
This robust two-step approach is made possible 
by modeling the relations using only the inner-product of the latent vectors
and introducing the orthogonal regularizer 
for preserving the encoded information.

\section{Experiments}
\label{sec:experiments}

\paragraph{Data}

To demonstrate the effectiveness of the proposed method,
we used the following two datasets: Wikipedia and Movielens.

The Wikipedia data were multilingual document-word relational datasets in
English (EN), German (DE), Italian (IT) and Japanese (JP),
where objects were documents and words,
and a document has a relation with a word 
when the document contained the word.
The documents were obtained from the following five categories in Wikipedia:
`Nobel laureates in Physics',
`Nobel laureates in Chemistry',
`American basketball players',
`American composers' and
`English footballers'.
For each category, we sampled 20 documents that appeared in all languages.
We used 1,000 frequent words after removing stop-words for each language.
There were 100 document objects and 1,000 word objects,
and 9,191 relations on average for each document-word
relational dataset in a language.

The Movielens data are a standard benchmark dataset 
for collaborative filtering~\cite{herlocker1999algorithmic}.
The original data contained 943 users, 1,682 movies, and 100,000 ratings.
First, we randomly split users into two sets.
Then, two user-item relational datasets were constructed 
by using the first or second sets of users and all items,
where objects were users and items, and a user and an item were connected
when the user had rated the item.
We call this data Movielens-User.
There were 471 user objects and 1,682 item objects in each dataset.
We also constructed Movielens-Item data
by randomly splitting items into two sets,
where there were 943 user objects and 841 item objects in each dataset.

For the evaluation measurements,
we used the top-$R$ accuracy, which is the rate that 
the correctly corresponding object is included 
in the top-$R$ list estimated with a method.

\paragraph{Comparing methods}

We compared the proposed method with the following five methods:
Degree, DeepWalk, CKS, Adversarial and ReMatch.
The Degree method finds correspondence between objects 
with its degree, or the number of relations of the object,
where objects with similar degree are considered as matched.

The DeepWalk method finds correspondence
with the Euclidean distance between 50-dimensional continuous feature vectors
obtained by the DeepWalk~\cite{perozzi2014deepwalk}.
The DeepWalk is a representation learning method for graphs 
and is not a method for matching, but we included it
as a comparing method since the proposed method is based on the DeepWalk.
The DeepWalk method estimates the feature vectors by maximizing the
likelihood (\ref{eq:likelihood}).
We used typical values for the DeepWalk hyperparameters as follows:
the number of walks that started from an object was 10,
the length of a walk was 80,
the window size of neighbors in a random walk sequence was 10,
the number of negative samples was 20,
the batch size was 4,096,
the Adam~\cite{kingma2014adam} with learning rate $10^{-2}$ was used
for optimization.

The CKS method is the convex kernelized sorting~\cite{djuric2012convex},
which is an unsupervised object matching method.
With the CKS, correspondences are found
by stochastically permuting objects so as 
to maximize the dependence between two
datasets, where the Hilbert Schmidt independence criterion (HSIC)
is used for measuring the dependence.
The ranking of objects to be matched is obtained by using 
the probability of a match estimated by the CKS.
The input of the CKS was the 50-dimensional continuous feature vectors
obtained by the DeepWalk method,
and the Gaussian kernels were used for calculating the HSIC.
We used the code provided by the authors~\cite{tuc}.
Note that with the CKS we found matching of documents (not for words) 
with the Wikipedia data,
users with the Movielens-User data, and items with the Movielens-Item data.

The Adversarial method orthogonally transforms the continuous feature vectors
obtained by the DeepWalk method onto a common latent space 
by matching the distributions 
using an adversarial approach~\cite{lample2018word,conneau2017word}.
With the Adversarial method, a neural network, 
which is called the discriminator, 
is trained so as to predict 
the dataset identifier where each transformed latent vector comes from.
The projection matrices are optimized so as to prevent the discriminator
to predict the dataset identifiers.
We used the code provided by the authors~\cite{muse},
and used the default hyperparameters.
The Euclidean distance between 50-dimensional transformed
continuous feature vectors was used for matching.

The ReMatch method is an unsupervised cluster matching method 
for relational data~\cite{iwata2016}. 
With the ReMatch, an object is assigned into a common cluster 
shared across all datasets 
based on stochastic block modeling~\cite{wang1987stochastic,kemp2006learning},
where 
the number of clusters is automatically estimated from the given data
using Dirichlet processes.
We considered that objects assigned in the same cluster were matched.

The proposed method trained with the following settings.
We obtained 50-dimensional continuous latent vectors
by the same setting with the DeepWalk method described above,
and transformed them by minimizing (\ref{eq:loss}).
For the kernel to calculate the MMD,
we used the following Gaussian kernel,
$k(\vec{u},\vec{u}')=\exp\left(-\frac{1}{2}\parallel\vec{u}-\vec{u}'\parallel^{2}\right)$.
We fixed the hyperparameter $\lambda=100$ in (\ref{eq:loss})
for all the datasets.
For optimization, 
we used the Adam~\cite{kingma2014adam} with the learning rate $10^{-2}$.
We ran the proposed method ten times with different initializations,
and the selected a result by the value of the loss function (\ref{eq:loss}).
The correspondence was found based on 
the Euclidean distance between
50-dimensional transformed continuous feature vectors.

\paragraph{Results}

\begin{figure}
\centering
{\tabcolsep=-0.5em
\begin{tabular}{ccc}
\includegraphics[width=0.37\linewidth]{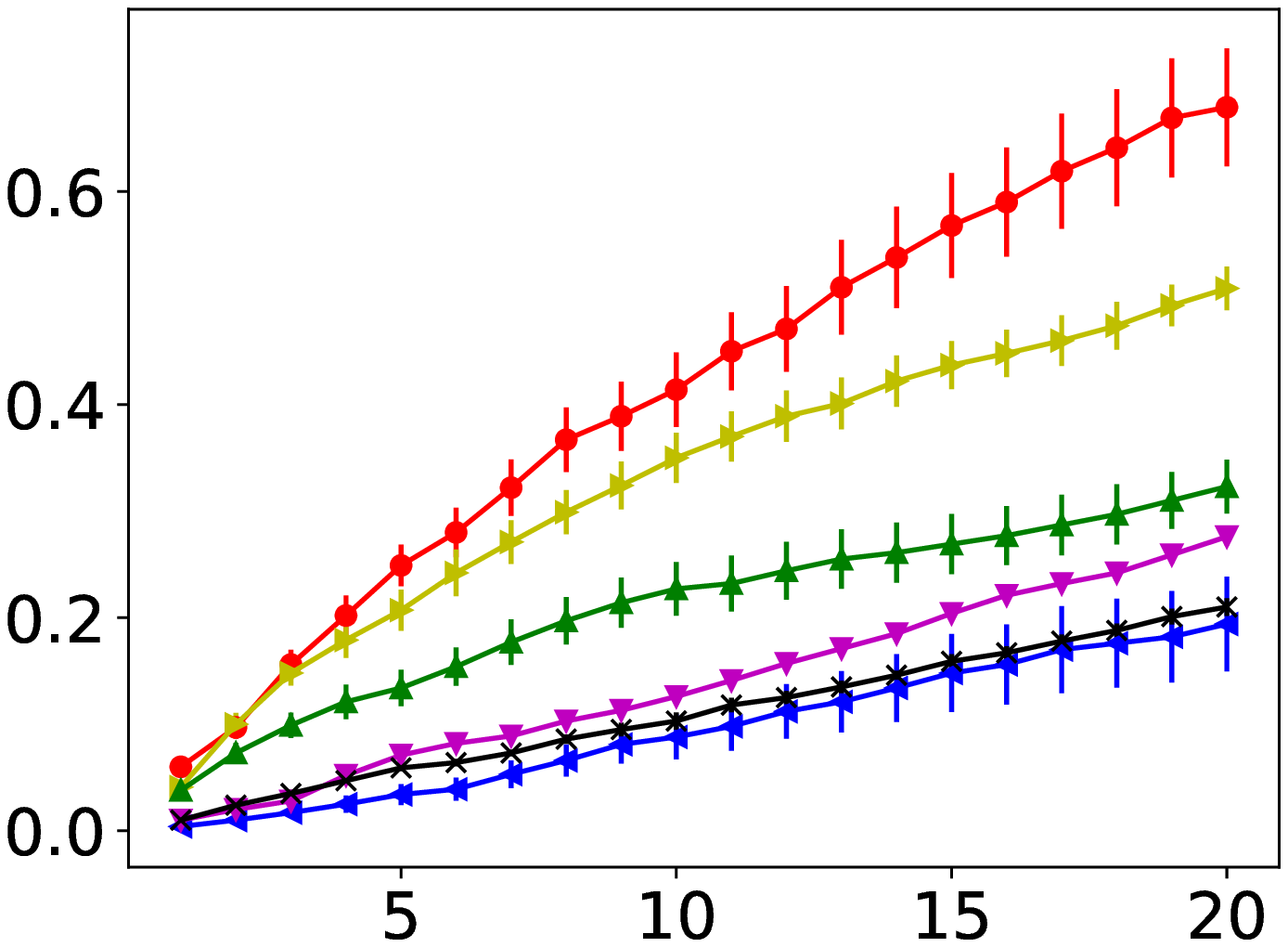} &
\includegraphics[width=0.37\linewidth]{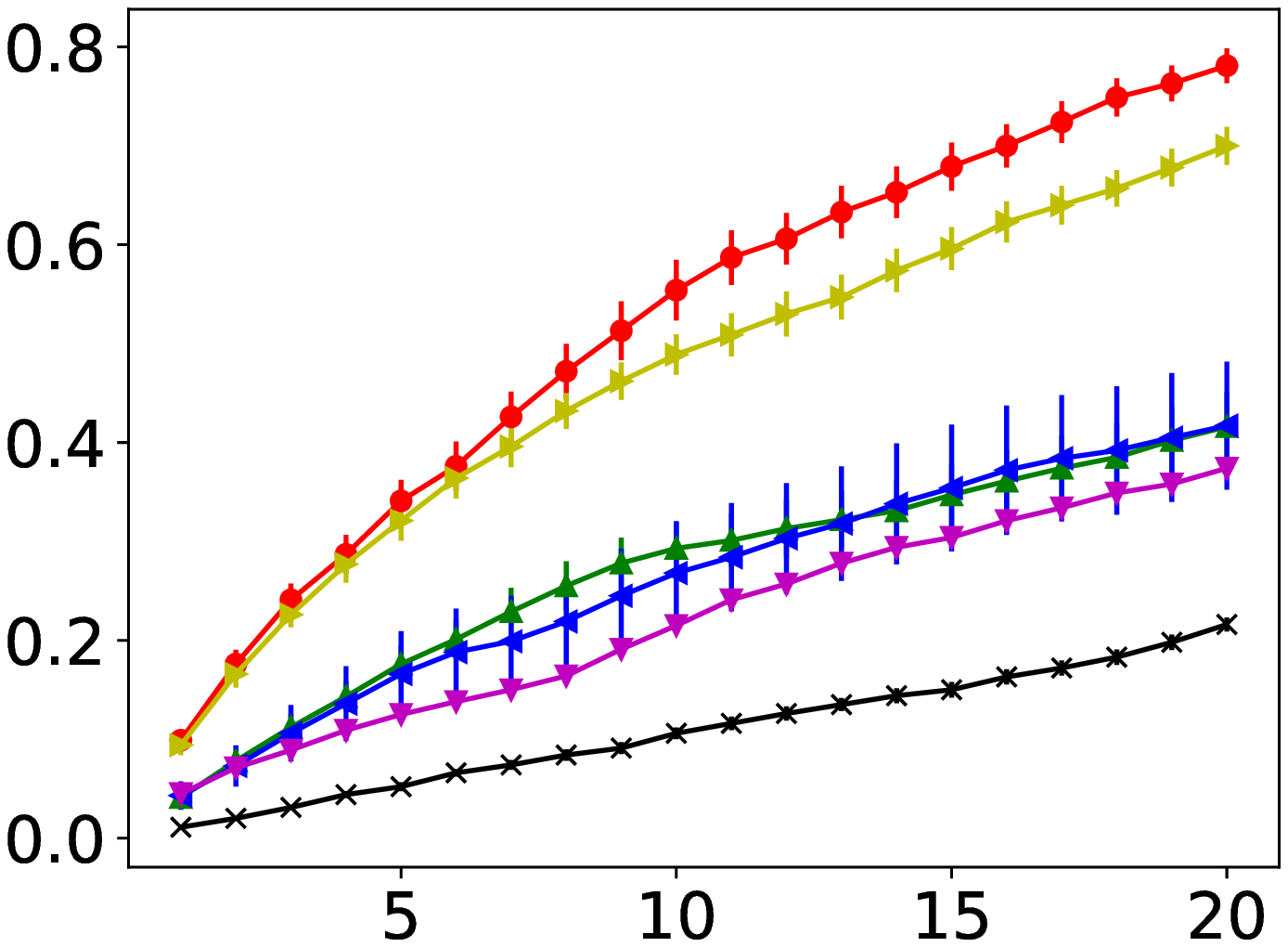} &
\includegraphics[width=0.37\linewidth]{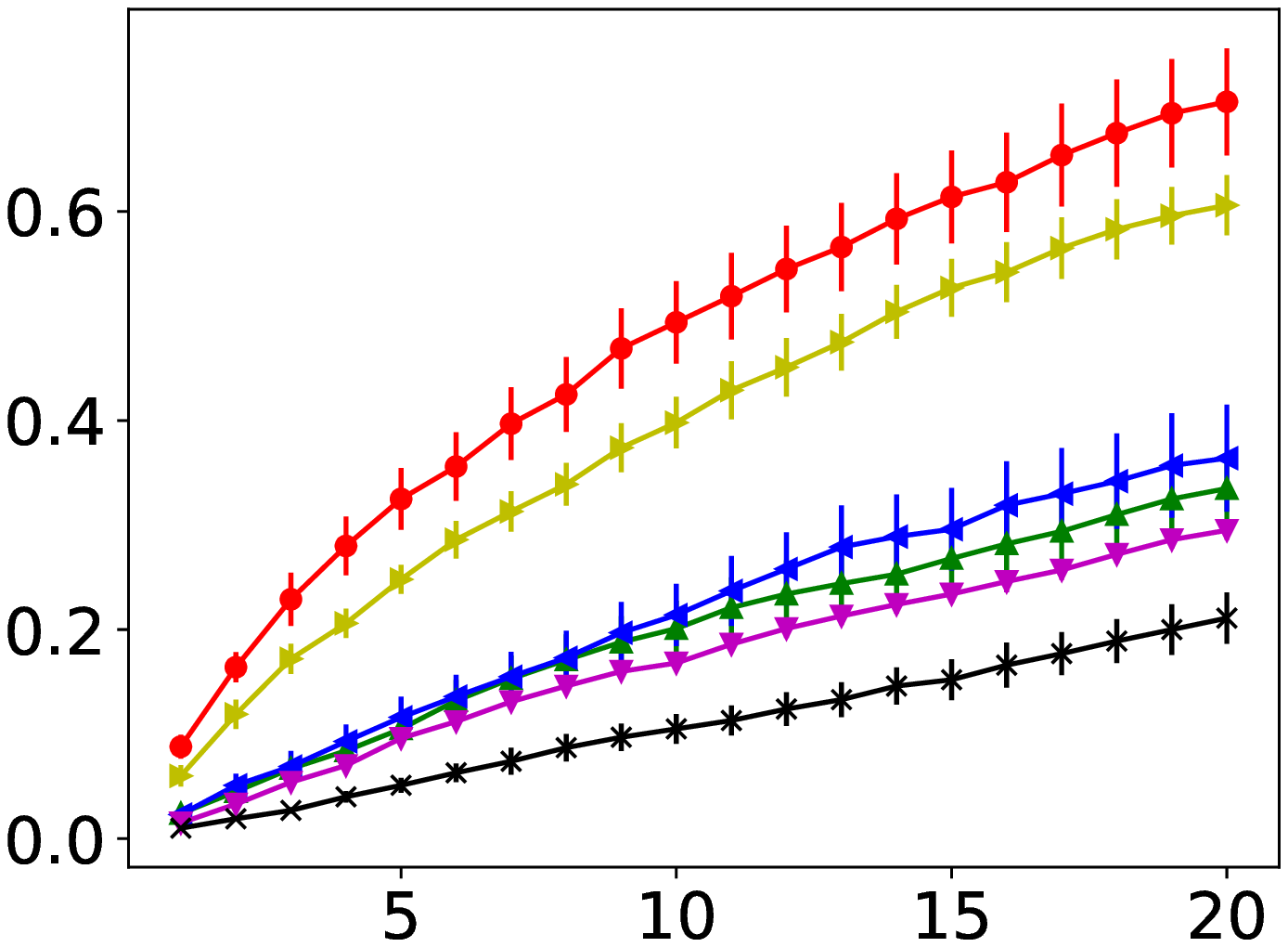}
\\
(a) Wikipedia-EN-DE & (b) Wikipedia-EN-IT & (c) Wikipedia-EN-JP \\
\includegraphics[width=0.37\linewidth]{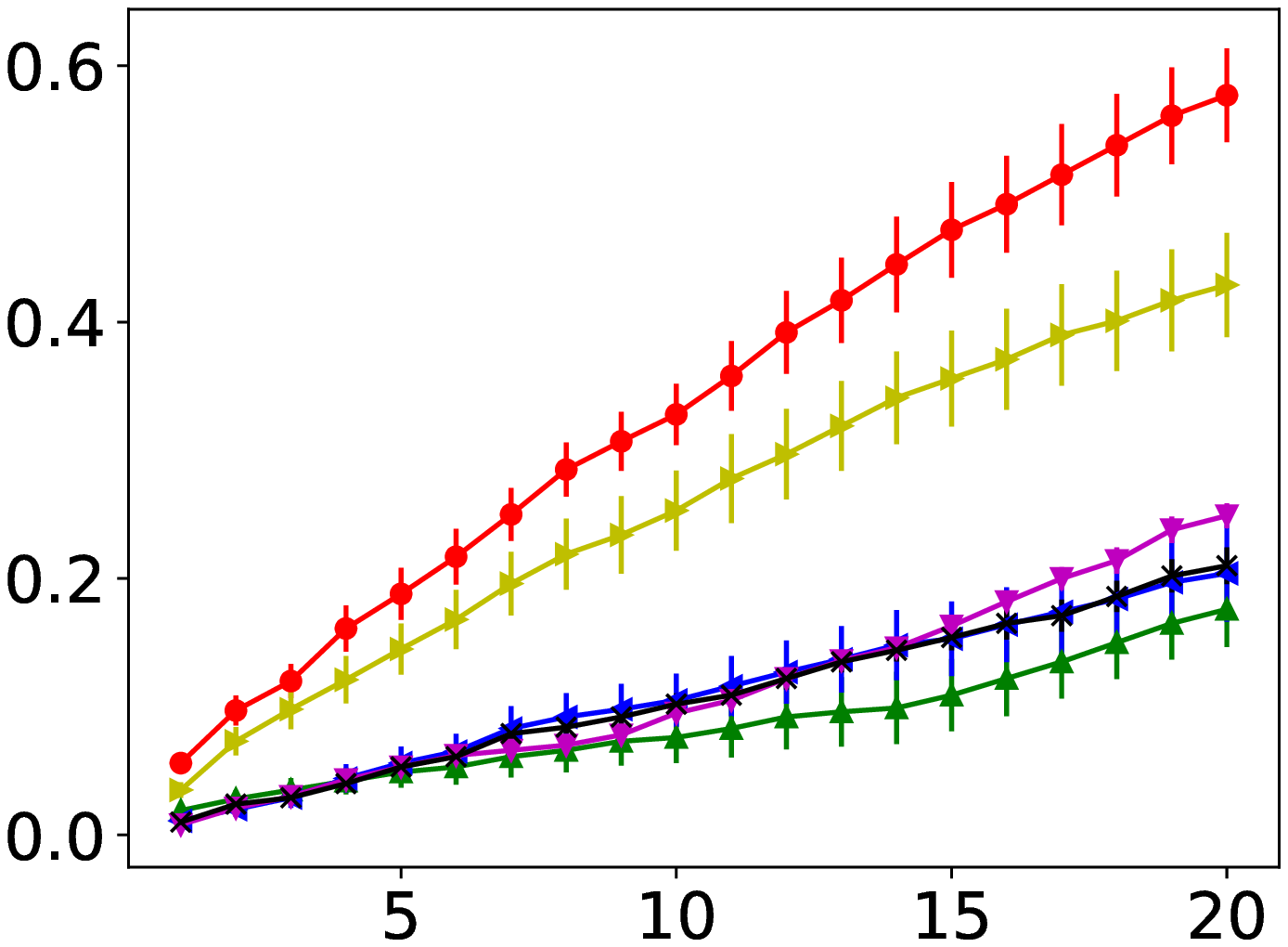} &
\includegraphics[width=0.37\linewidth]{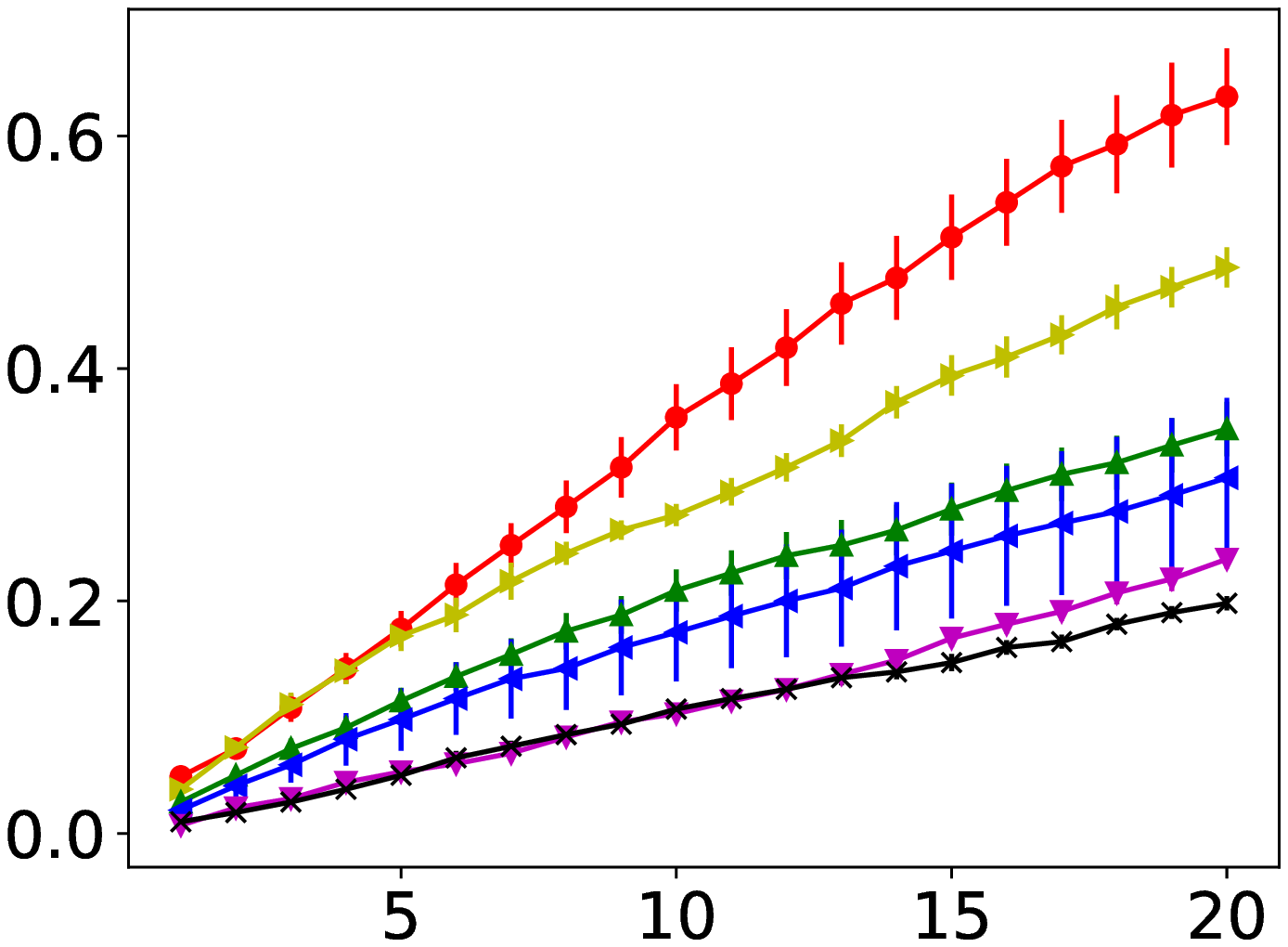} &
\includegraphics[width=0.37\linewidth]{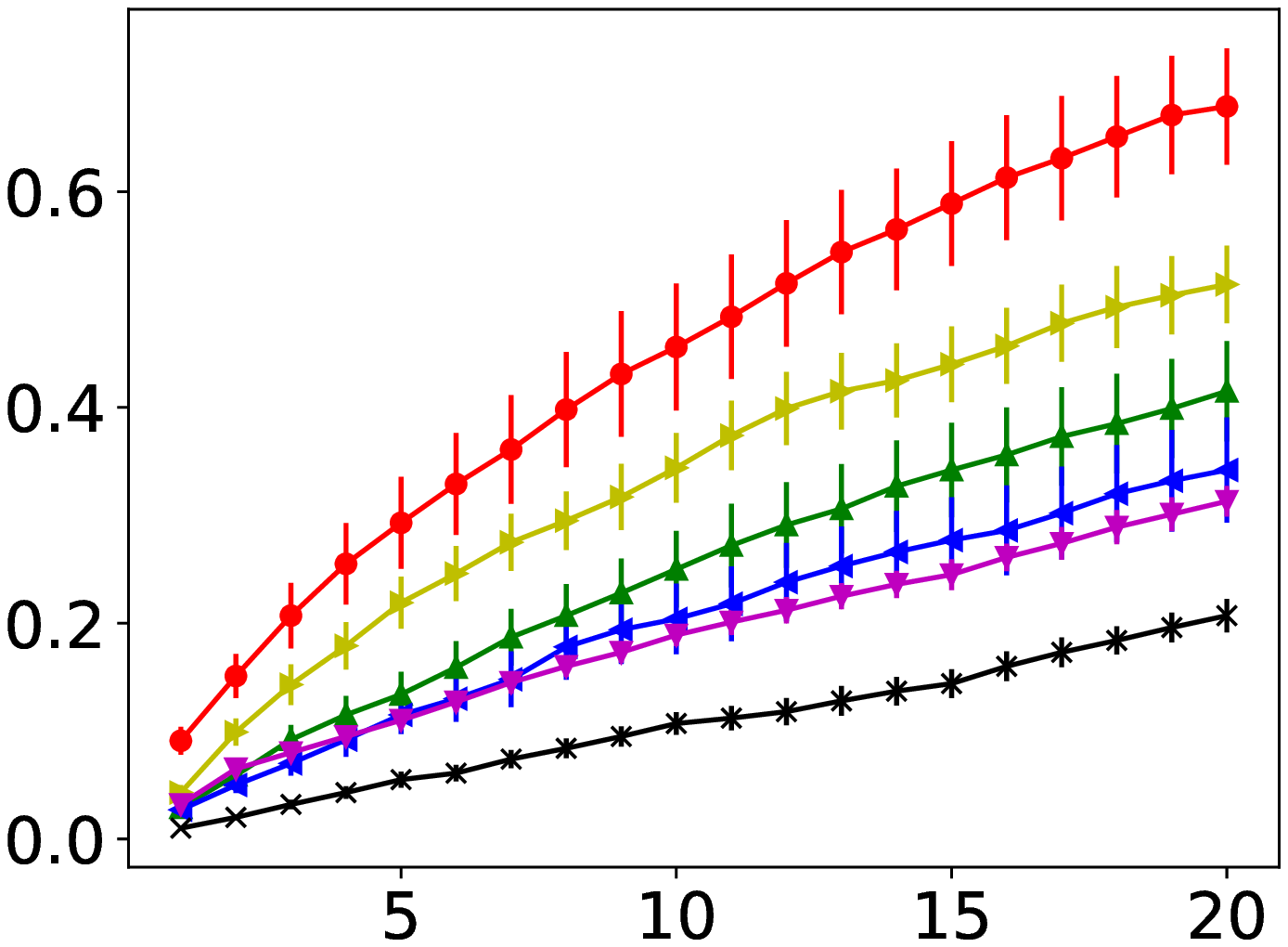}
\\
(d) Wikipedia-DE-IT & (e) Wikipedia-DE-JP & (f) Wikipedia-IT-JP \\
\includegraphics[width=0.37\linewidth]{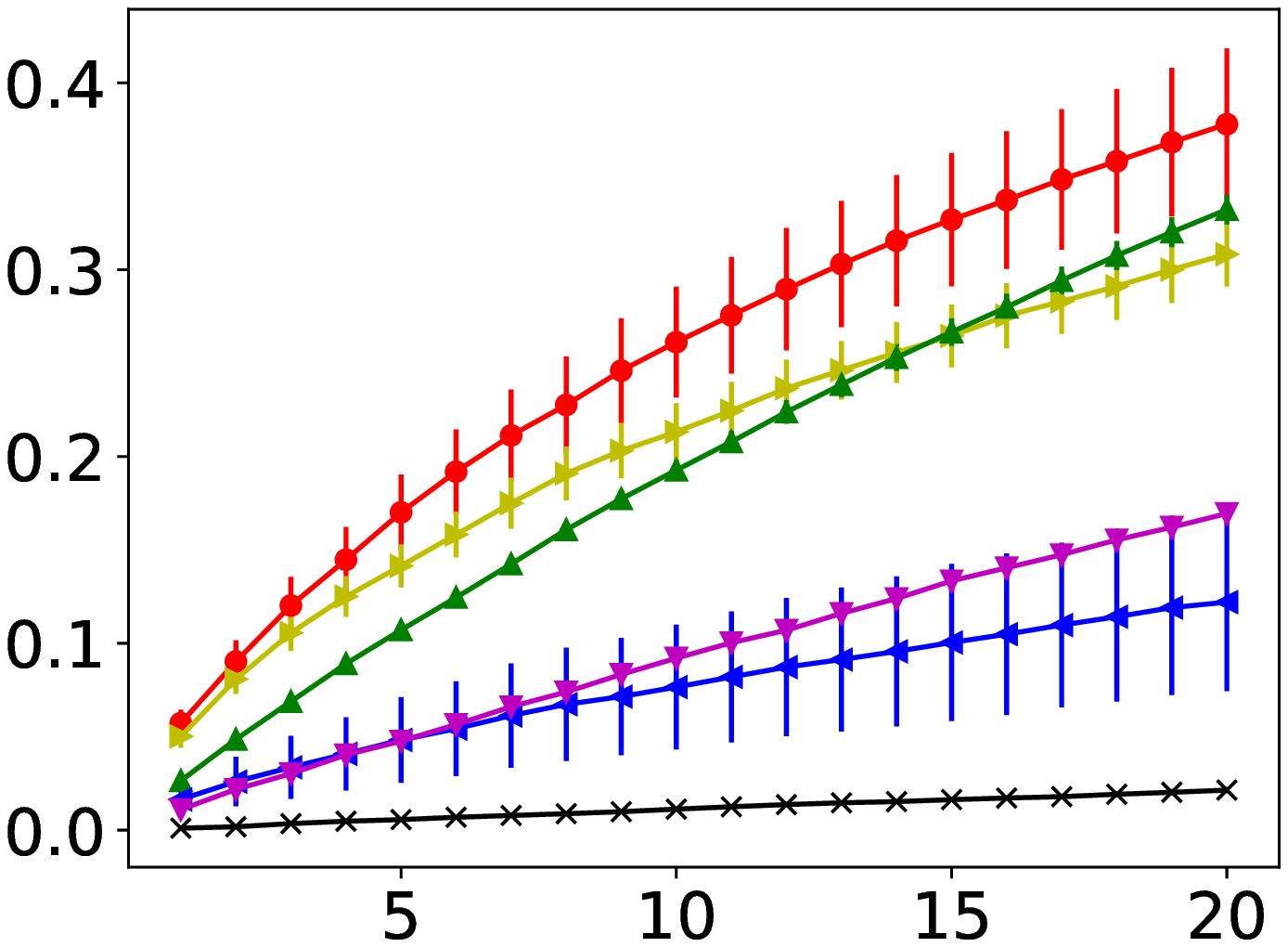} &
\includegraphics[width=0.37\linewidth]{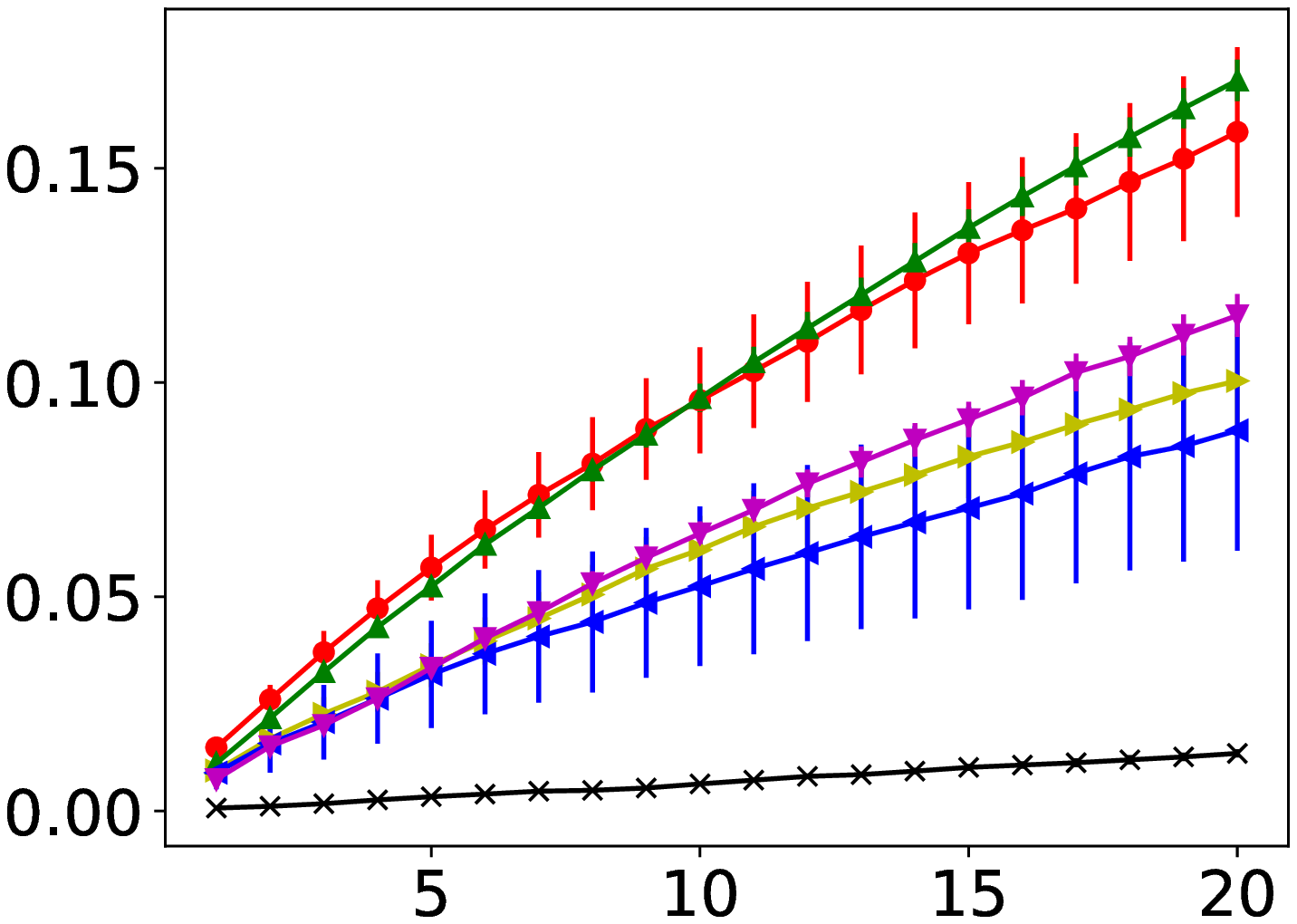} &
\includegraphics[width=0.37\linewidth]{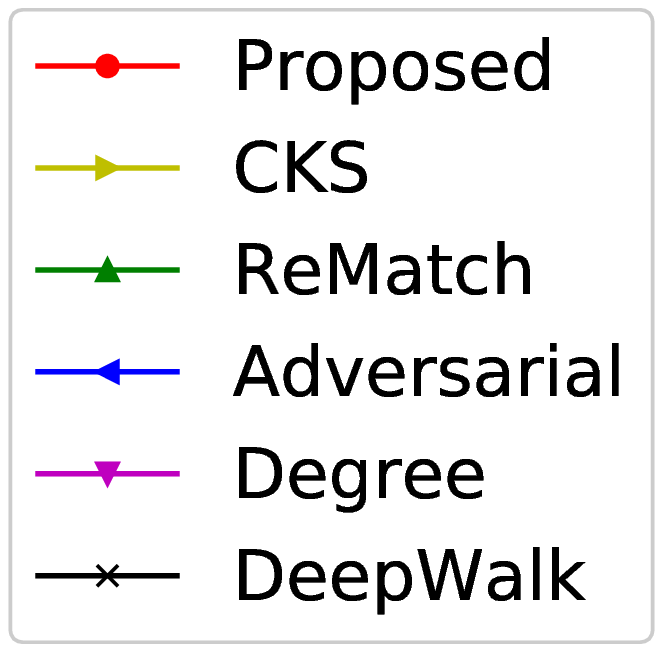}
\\
(g) Movielens-User & (h) Movielens-Item & \\
\end{tabular}}
\caption{Top-$R$ accuracy with Wikipedia (a-f) and Movielens (g,h) datasets. The x-axis is $R$, the y-axis the average top-$R$ accuracy, and the errorbar shows the standard error.}
\label{fig:accuracy}
\end{figure}

Figure~\ref{fig:accuracy} shows the top-$R$ accuracy with the Wikipedia and Movielens datasets, which are averaged over ten experiments.
The proposed method achieved the highest accuracy in seven
of the eight datasets,
and the second highest accuracy in the Movielens-Item data (h).
The accuracy of the DeepWalk method was low, 
which was almost the same with random matching.
This was reasonable because the DeepWalk obtained the latent vectors 
for each dataset independently. 
The CKS, Adversarial and the proposed methods 
used the independent latent vectors estimated by the DeepWalk as inputs,
and achieved the higher accuracy than the DeepWalk.
This result indicates that these methods found the relationship
between datasets in an unsupervised fashion.
The performance of the CKS was worse than the proposed method.
It would be because the CKS finds alignments 
based on kernel matrices of the latent vectors without 
using their characteristics.
On the other hand, the proposed method exploits their characteristics,
where the structural information is encoded as their inner-products,
as the orthogonality regularization.
The Adversarial method gave the lower accuracy
compared with the proposed method.
Note that we did not tune the hyperparameters of the Adversarial method,
and careful hyperparameter tuning would give better performance.
However, it is well known
that the adversarial training often
becomes unstable~\cite{arjovsky2017towards}, 
and the hyperparameter tuning is difficult. 
In contrast, the proposed method does not need additional neural networks,
i.e. a discriminator in the Adversarial method,
and the distributions are matched stably by using the MMD.
Since the ReMatch represents an object by its cluster assignment,
some objects in the same cluster have the identical representation,
and the accuracy of the ReMatch was low.

\begin{figure}[t]
\centering
{\tabcolsep=-0.5em
\begin{tabular}{ccc}
\includegraphics[width=14.5em,height=14.5em]{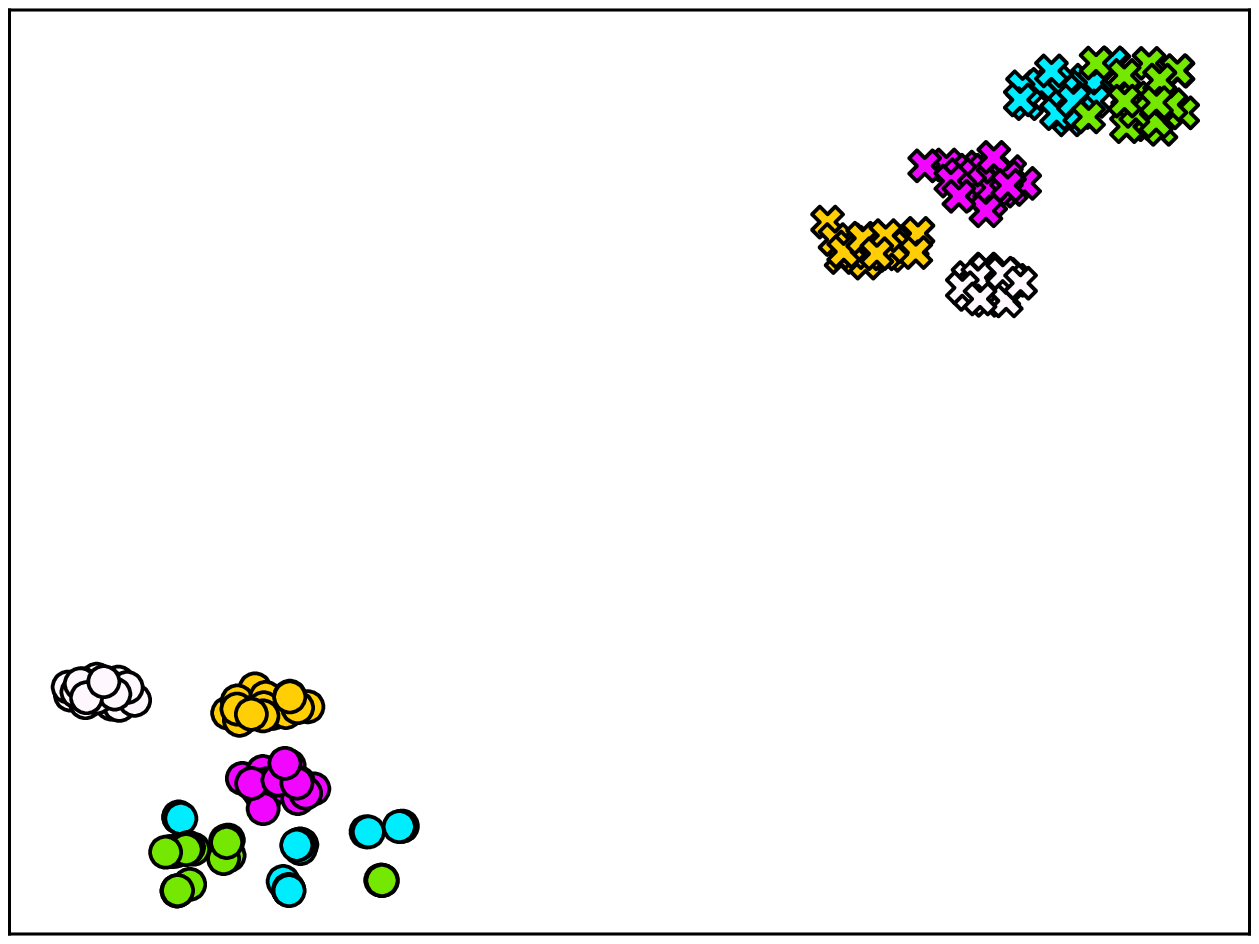}
&
\includegraphics[width=14.5em,height=14.5em]{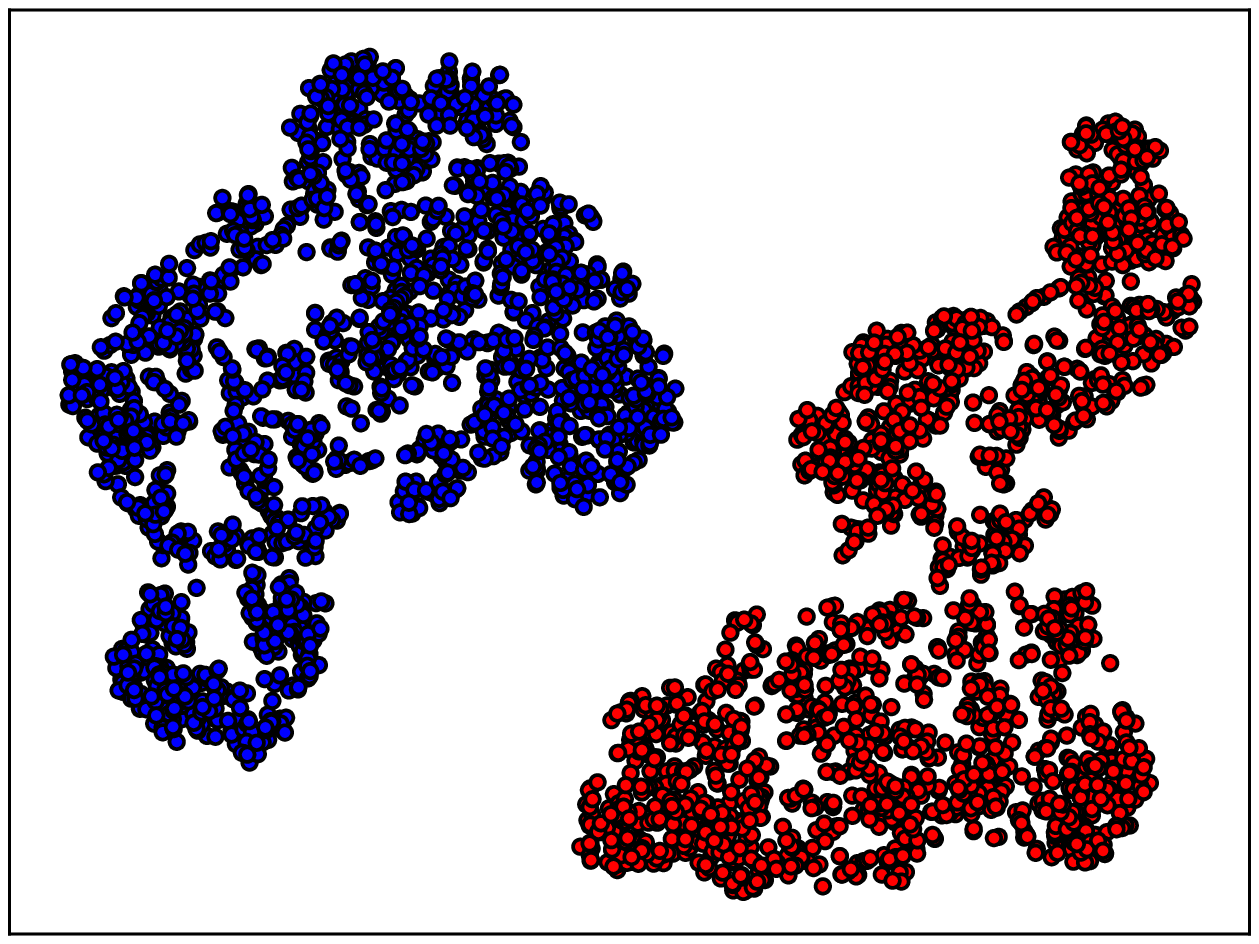}
&
\includegraphics[width=14.5em,height=14.5em]{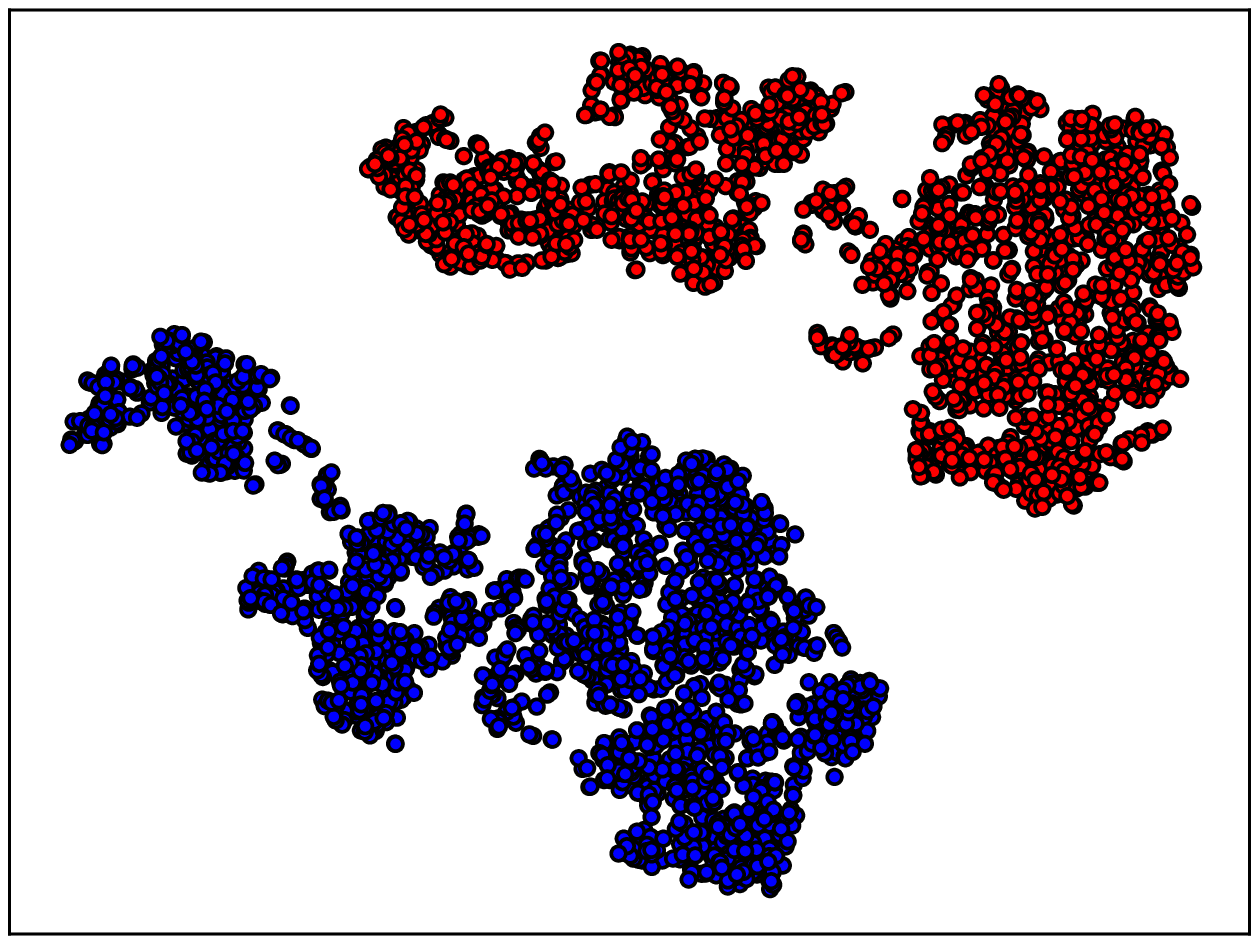}
\\
\multicolumn{3}{c}{Latent vectors with the DeepWalk method}\\
(a) Wikipedia-EN-DE &
(b) Movielens-User &
(c) Movielens-Item \\
\includegraphics[width=14.5em,height=14.5em]{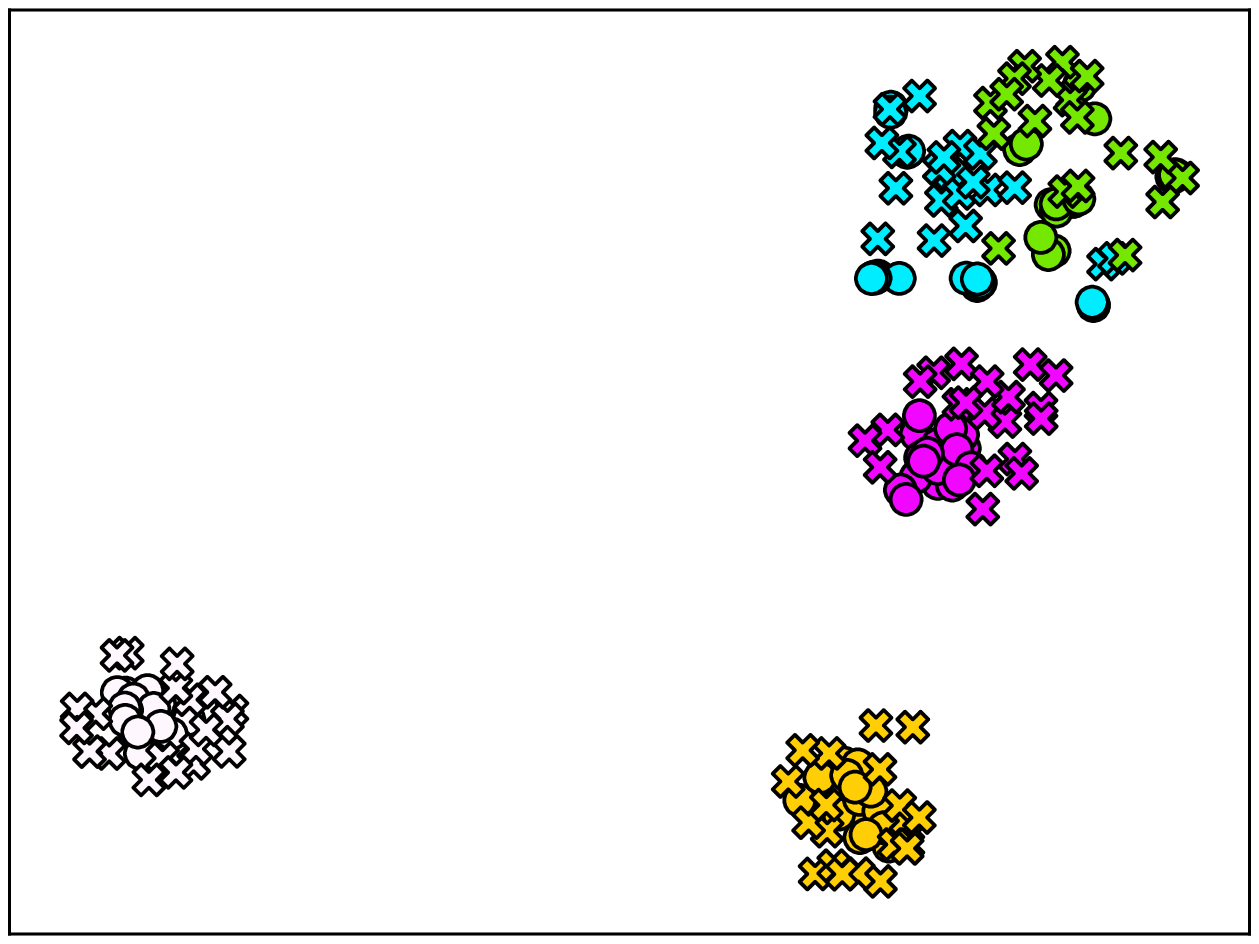}
&
\includegraphics[width=14.5em,height=14.5em]{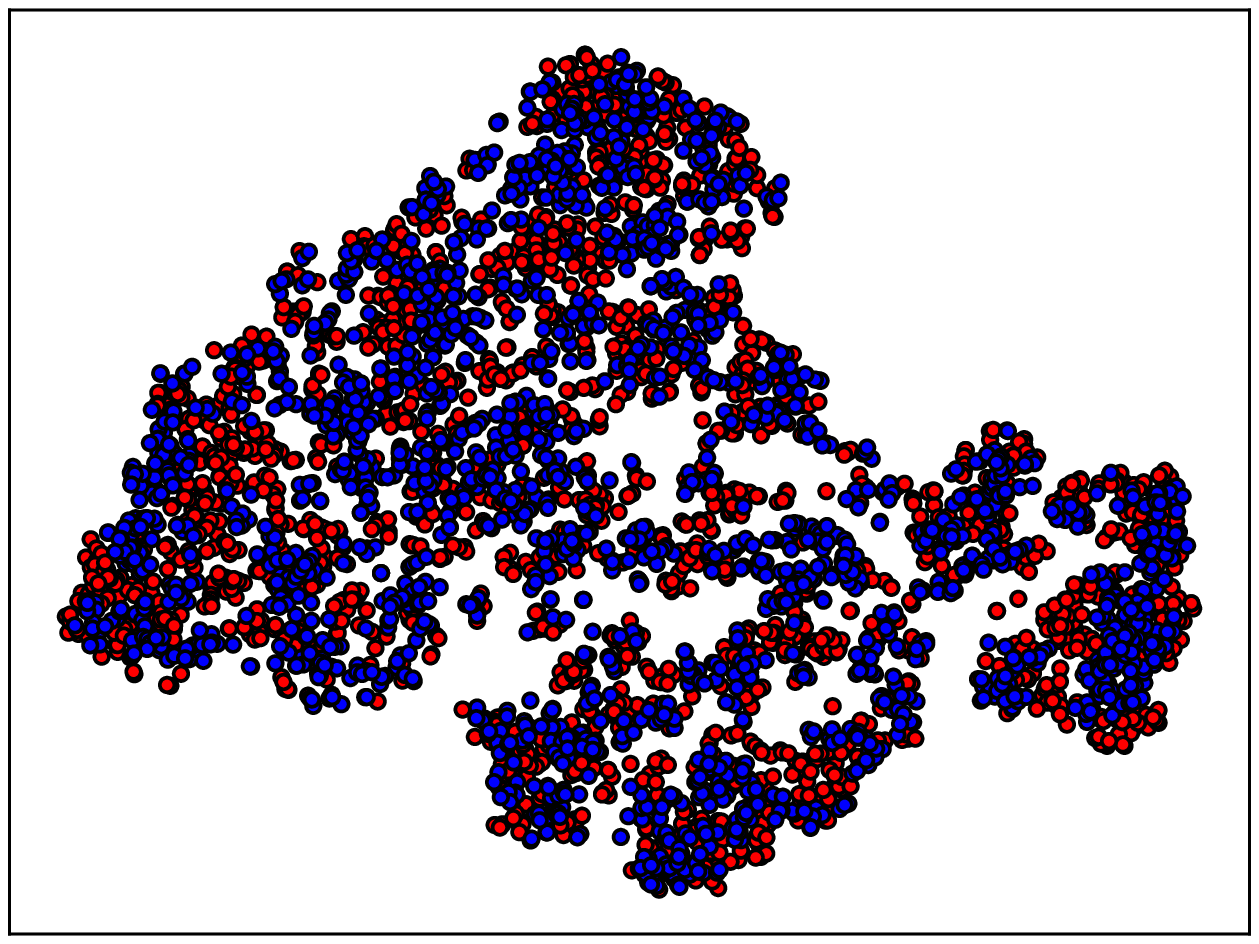}
&
\includegraphics[width=14.5em,height=14.5em]{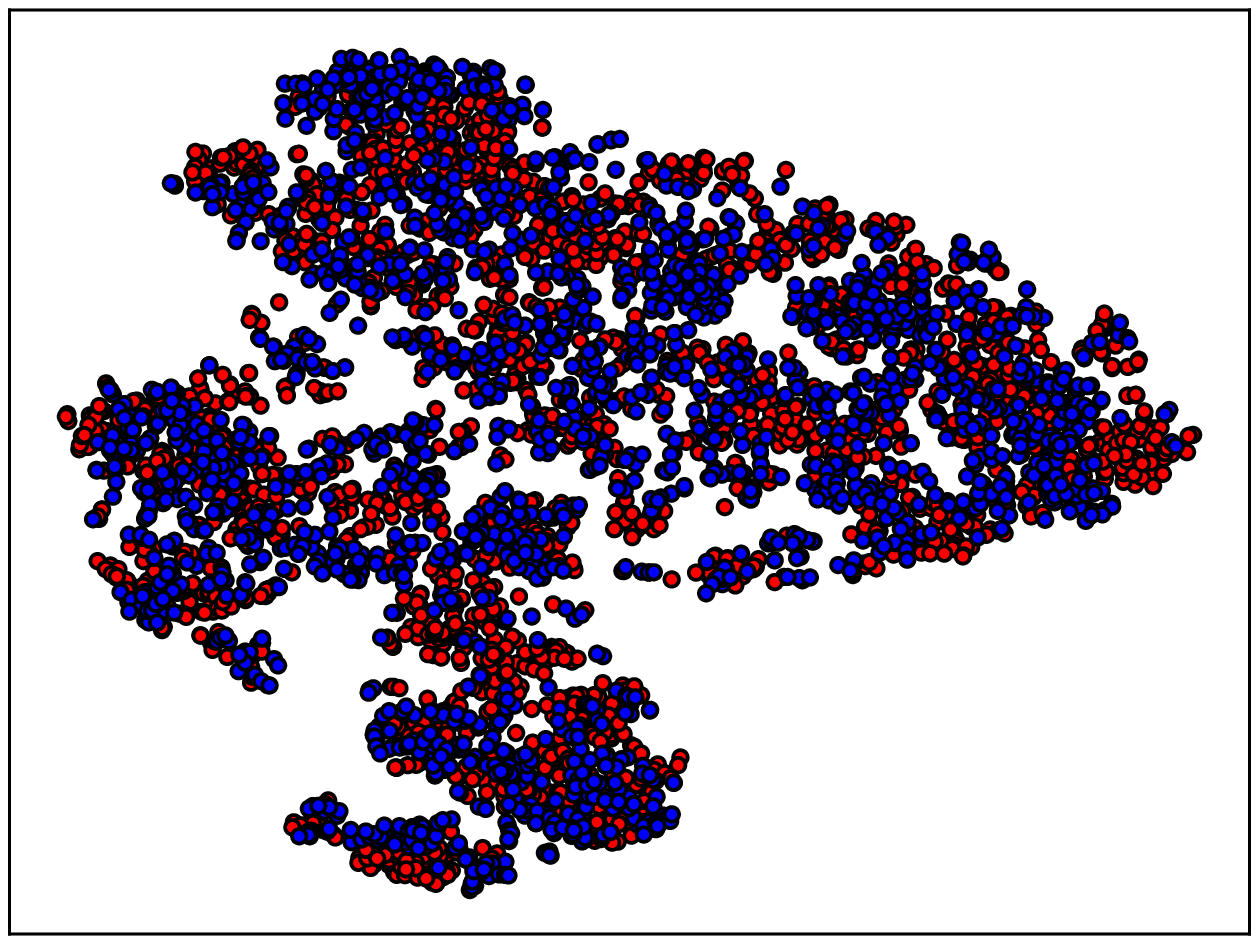}
\\
\multicolumn{3}{c}{Transformed latent vectors with the proposed method}\\
(d) Wikipedia-EN-DE &
(e) Movielens-User &
(f) Movielens-Item \\
\end{tabular}}
\caption{The two-dimensional visualization of the latent vectors with $t$-SNE. The upper row shows the latent vectors with the DeepWalk method. The lower row shows the transformed latent vectors with the proposed method. (a,d) Wikipedia document-word relational datasets in English and German. Only document objects are shown, where the marker shape represents languages (`o': English, `x': German), and the marker color represents categories (blue: Nobel laureates in Physics, green: Nobel laureates in Chemistry, yellow: American basketball players, magenta: American composers, white: English footballers). (b,e) Movielens-User data. Only user objects are shown, where the marker color represents the dataset identifier (red: dataset1, blue: dataset2). (c,f) Movielens-Item data. Only item objects are shown, where the marker color represents the dataset identifier(red: dataset1, blue: dataset2).}
\label{fig:vis}
\end{figure}

Figure~\ref{fig:vis} shows 
the two-dimensional visualization of the latent vectors
with $t$-stochastic neighbor embedding ($t$-SNE)~\cite{maaten2008visualizing}
before (a,b,c) and after (d,e,f) the transformation.
The $t$-SNE is a nonlinear dimensionality reduction method.
With the Wikipedia data, the DeepWalk obtained similar latent vectors for 
documents in the same category in each language,
but obtained dissimilar latent vectors for different languages (a).
The proposed method successfully found the category alignment,
which was shown in (d), where the documents in the same category 
in different languages were located close together.
Similarly, the latent vectors by the DeepWalk with the Movielens data (b,c)
were transformed so as to matching the distributions
by the proposed method (e,f).

Table~\ref{tab:time} shows the computational time in seconds.
The computational time with the CKS method 
increased cubically with the number of objects,
and did not scale to large data.
On the other hand, the proposed, DeepWalk and Adversarial methods
scale well by using stochastic gradient methods.

\begin{table}[t]
\centering
\caption{Computational time in seconds.}
\label{tab:time}
\begin{tabular}{lrrrrr}
\hline 
& Proposed & DeepWalk & Adversarial & CKS & ReMatch \\
\hline
Wikipedia & 1,204 & 291 & 3,153 & 317 & 297 \\
Movielens-User & 4,473 & 394 & 3,268 & 17,764 & 679 \\
Movielens-Item & 5,337 & 378 & 3,299 & 103,950 & 651 \\
\hline
 \end{tabular}
\end{table}

\section{Conclusion}
\label{sec:conclusion}

We proposed an unsupervised object matching methods for relational data.
With the proposed method,
object representations that contain hidden structural information
are obtained for each relational dataset,
and the representations are transformed onto a common latent space shared across all datasets
by matching the distributions while preserving the structural information.
With the experiments,
we confirmed that the proposed method can
effectively and efficiently find matchings in relational data.
For future work,
we would like to extend the proposed method for a semi-supervised setting, 
where a few correspondences are given.


\bibliographystyle{abbrv}

\end{document}